# Possibilistic Pertinence Feedback and Semantic Networks for Goal's Extraction

Mohamed Nazih Omri

*Department of Technology*
*Preparatory Institute of Engineering Studies of Monastir*
*Kairouan Road, 5019 Monastir, Tunisia*
E-mail:nazih.omri@ipeim.rnu.tn

## Abstract

Pertinence Feedback is a technique that enables a user to interactively express his information requirement by modifying his original query formulation with further information. This information is provided by explicitly confirming the pertinent of some indicating objects and/or goals extracted by the system. Obviously the user cannot mark objects and/or goals as pertinent until some are extracted, so the first search has to be initiated by a query and the initial query specification has to be good enough to pick out some pertinent objects and/or goals from the Semantic Network. In this paper we present a short survey of fuzzy and Semantic approaches to Knowledge Extraction. The goal of such approaches is to define flexible Knowledge Extraction Systems able to deal with the inherent vagueness and uncertainty of the Extraction process. It has long been recognised that interactivity improves the effectiveness of Knowledge Extraction systems. Novice user's queries are the most natural and interactive medium of communication and recent progress in recognition is making it possible to build systems that interact with the user. However, given the typical novice user's queries submitted to Knowledge Extraction Systems, it is easy to imagine that the effects of goal recognition errors in novice user's queries must be severely destructive on the system's effectiveness. The experimental work reported in this paper shows that the use of possibility theory in classical Knowledge Extraction techniques for novice user's query processing is more robust than the use of the probability theory. Moreover, both possibilistic and probabilistic pertinence feedback can be effectively employed to improve the effectiveness of novice user's query processing.

## 1. Introduction

The effectiveness of a KES is therefore crucially related to the system's capability to deal with the vagueness and uncertainty of the Extraction process. Commercially available KESs generally ignore these aspects; they oversimplify both the representation of the Objects' content and the user-system interaction. The goal of a Knowledge Extraction System (KES) is to extract knowledge considered pertinent to a user's request, expressed in the natural language. The pertinence feedback of a KES is measured through parameters, which reflect the ability of the system to accomplish such goal. However, the nature of the goal is not deterministic, since uncertainty and vagueness are present in many different parts of the extraction process. The user's expression of his/her knowledge needs in a request is uncertain and often vague, the representation of an Object and/or a Goal informative content is uncertain, and so is the process by which a request representation is matched to an Object representation.

The research in KE has aimed at modeling the vagueness and uncertainty, which invariably characterize the management of knowledge. A first glens of approaches is based on methods of analysis of natural language [1]. The main limitation of these methods is the level of deepness of the analysis of the language, and their consequent range of applicability: a satisfying interpretation of the Objects' meaning needs a too large number of decision rules even in narrow application domains. A second glens of approaches is more general: their objective is to define Extraction models, which deal with imprecision, and uncertainty independently on the application domain. The set of approaches belonging to this class goes under the name of Probabilistic Information Retrieval (IR) [2]. There is another set of approaches receiving increasing interest that aims at applying techniques for dealing with vagueness and uncertainty. This set of approaches goes under the name of *Knowledge Extraction (KE).*

The results of an experimental study of the effects of the use of possibilistic and probabilistic technics in novice user's queries on the effectiveness of a KE system are presented in this paper. The paper is structured as follows. Section 2 describes the structure of fuzzy Goals and fuzzy Objects used in the Semantic Network. Section 3 presents an introduction to the KES problem. This section also gives the motivations of the work reported here.

Section 4 reports on the use of probabilistic relevance feedback as a first way to improve the effectiveness of the User's Query Processing task. Section 5 reports on the use of possibilistic pertinent feedback as a second way to improve the effectiveness of the User's Query Processing task. Section 6 presents the experimental environment of the study: the test collection, the queries, and the KE system used. The limitations of the study are reported in 7. In section 8 we draw the conclusions of our study.

## 2. Fuzzy Goals and Fuzzy Objects

In possibility theory, there has been a lot of research into the notion of Knowledge Extraction. However, researchers have mainly analysed the information retreival from document, while we need a somewhat more sophisticated notion: Goal Extraction from fuzzy sets to fuzzy goals in a Semantic Network. To extend definitions of Knowledge (goal) Extraction from fuzzy sets to fuzzy goals, we must first analyse how fuzzy objects and goals can be described in terms of fuzzy sets [3, 4].

How does a user formulate his or her gaols? For example, how do we describe a goal when we look for a house to buy? A natural goal is to have a car not too old, not too expensive, a red color, etc. In general, to describe a goals:

- we list *attributes* (in the above example, age, cost, and color), and
- we list the desired (fuzzy) value $A_1,…,A_n$ of these attributes (in the above example, these values are, correspondingly, "not too old", "not too expensive", and "red").

Each of the fuzzy values like "not too old" can be represented, in a natural way, as a fuzzy set.
Similarly, an object can be described if we list the attributes and corresponding values. At first glance, it may seem that from this viewpoint, a description of an object is very much alike the description of a goal, but there is a difference.

## 3. Knowledge Extraction System

In this paper we will survey the application to KE of three theories that have been used in Artificial Intelligence for quite some time: possibility, probability ans SN theories. The use of these techniques in KE has been recently refered to as Knowledge Extraction in analogy with the areas called Computing and Information Retreival[5, 6].

*Fuzzy set theory* [7] is a formal framework well suited to model vagueness: in information retreival it has been successfully employed at several levels [8, 9], in particular for the definition of a superstructure of the Boolean model[10], with the appealing consequence that Boolean KESs can be improved without redesigning them completely [11, 12, 13]. Through these extensions the gradual nature of pertinence of Objects to user's requests can be modelled.

A Knowledge Extraction system is a computing tool, which represents and stores knowledge to be automatically extracted for future use. Most actual KE systems store and enable the Extraction of only knowledge or Objects. However, this is not an easy task, it must be noticed that often the sets of Objects a KES has to deal with contain several thousands or sometimes millions of Objects and/or Goals.

A user accesses the KES by submitting a request; the KES then tries to extract all Objects and/or Goals that are "pertinent" to the request. To this purpose, in a preliminary phase, the Objects contained in the Semantic Network (SN) [14, 15, 16] are analyzed to provide a formal representation of their contents: this process is known as "indexing"[4]. Once an Object has been analyzed a surrogate describing the Object is stored in an index, while the Object itself is also stored in the SN. To express some knowledge needs a user formulates a request, in the system's request language. The request is matched against entries in the index in order to determine which Objectsand/or goals are pertinent to the user. In response to a request, a KES can provide either an exact answer or a ranking of Objects that appear likely to contain knowledge pertinent to the request. The result depends on the formal model adopted by the system. In our KES, requests are expressed in natural language.

A different approach is based on the application of the SN *theory* to KE. Semantic Networks have been used in this context to design and implement KESs that are able to adapt to the characteristics of the KE environment, and in particular to the user's interpretation of pertinent. In this chapter we will review the applications of fuzzy set theory and Semantic networks to KE.

## 4. Probabilistic Relevance Feedback

*Probabilistic relevance feedback* (PRF) is a technique that consists of adding new goals to the original query. The goals added are chosen by taking the first m goals in a list where all the goals present in pertinent objects are ranked according to the following weighting function [17]:

$$w_i = \log \frac{r_i(N - n_i - R + r_i)}{(R - r_i)(n_i - r_i)} \qquad (1)$$

where: N is the number of objects in the collection, $n_i$ is the number of objects with an occurrence of goal i, R is the number of pertinent objects pointed out by the user, and $r_i$ is the number of pertinent objects pointed out by the user with an occurrence of goal *i*. Essentially, PRF compares the frequency of occurrence of a goal in the objects marked as pertinent with its frequency of occurrence in the whole object collection. If a goal occurs more frequently in the objects marked as pertinent as in the whole object collection it is assigned a higher weight. There are two ways of choosing the goals to add to the query:

- adding goals whose weight is over a predefined threshold,
- or adding a fix number of goals, the k goals with the highest weight.

In the experiments reported in this paper we used the second technique. After a few tests, the number of terms added to the original query was set to 10. This number has been proved experimentally to be quite effective, without modifying the extending the scope of the query too much [17].

## 5. Possibilistic Pertinence Feedback

Let, $E_1$, $E_2$... $E_p$ a subsets no emptiness, two to two distinct, of $\Omega$ (supposed finished), where carry weights of probability, noted $m(E_1)$... $m(E_p)$, as, according to Shafer[18]:

$$\sum_{i=1,p} m(E_i) = 1, \qquad (2)$$

and

$$\forall \, i, m(E_i) \succ 0 \qquad (3)$$

In this situation, the probability of an event TO will be imprecise that to be-to-say restrained in an interval $[P_*(A), P^*(A)]$, defined by :

$$P_*(A) = \sum_{E_i \subseteq A} m(E_i) = \sum_i m(E_i) N_{E_i}(A) \qquad (4)$$

$$P^*(A) = \sum_{E_i \subseteq A \neq \emptyset} m(E_i) = \sum_i m(E_i) \Pi_{E_i}(A) \qquad (5)$$

We verify that the function P* is a measure of possibility if and only if $E_1 \subset E_2 \subset \ldots \subset E_p$ then the possibility distribution $\pi$ associated to P* and $P_*$ is defined by Dubois-Prade[19]:

$$\forall w, \pi(w) = P^*(\{w\}) = \sum_{j=i}^{p} m(E_j) \quad if \quad w \in E_i, w \notin E_{i-1} \\ = 0 \quad if \quad w \in \Omega - E_p \qquad (6)$$

We then define the possibilistic inverse object frequency ($\pi_{iof}$) formula given by:

$$\pi_{iof}(g_i) = P_*(\{g_i\}) = -\sum_{j=1}^{p} \log \frac{n_i}{N} \qquad (7)$$

where $n_i$ is the number of objects in which the goal $g_i$ occurs, and $N$ is the total number of objects in the SN.

The possibilistic goal frequency ($\pi_{gf}$) is defined as:

$$\pi_{gf}(g_i, o_j) = \frac{\log(f_{i,j} + 1)}{\log(L_j)} \qquad (8)$$

where $f_{i,j}$ is the possibilistic frequency of goal $g_i$ in object $o_j$, and $L_j$ is the number of unique goals in object $o_j$.

We then define a possibilistic score $\pi_S$ of each object by summing the $\pi_{gf}$ and $\pi_{iof}$ possibilistic weights of all query goals found in the object:

$$\pi_s(o_j, q) = \sum_{g_i \in q} \pi_{iof}(g_i) \pi_{gf}(i, j) \qquad (9)$$

In the KE literature there exist many variations of this formula dependeng on the way of $\pi_{gf}$ and $\pi_{iof}$ weights are computed[11, 13]. We chose this one because it is the most standard schema.

A technique that enables a user to interactively express his goal requirement by modifying his original query formulation with further information is the standard Relevance Feedbak[14, 20, 21]. This additional information is provided by explicitly confirming the pertinent of some indicating object extracted by the system. Obviously the user cannot mark objects as pertinent until some are extracted, so the first search has to be initiated by a query and the initial query specification has to be good enough to pick out some pertinent objects from the SN. The user can mark the object(s) as pertinent and starts the PF process.

Among the many algorithms for PF, Probabilistic Relevance Feedback was chosen and implemented. Briefly, Probabilistic PF consists of adding new goals to the original query. The goals added are chosen by taking the first $k$ goals in a list where all the goals present in pertinent objects are ranked according to the following pertinent weighting function [13]:

$$pw(g_i) = r_i \cdot \log \frac{(r_i + 0.5)(N - n_i - R + r_i + 0.5)}{(R - r_i + 0.5)(n_i - r_i + 0.5)} \qquad (10)$$

where: N is the number of objects in the semantic network, $n_i$ is the number of objects with at least one occurrence of goal $g_i$, R is the number of pertinent objects used in the PF, and $r_i$ is the number of pertinent objects in R with at least one occurrence of goal $g_i$. The possibilistic score $\pi_S$ for each object is then calculated using the following formula that uses the pertinent weight $pw(g_i)$ of the goal $g_i$ instead of $\pi_{iof}(g_i)$.

$$\pi_{S\_RF}(o_j, q) = \sum_{g_i \in q} \pi_{pw}(g_i) \pi_{pgf}(i, j) \qquad (11)$$

Possibilistic Pertinence Feedback compares the possibility of occurrence of a goal in the objects marked as pertinent with its possibility of occurrence in the whole object in the SN. If a goal occurs more possible in the objects marked as pertinent than in the whole object in the SN it is assigned a higher $pw(g_i)$ weight. Then, there are two ways of choosing the goals to add to the query:

- adding goals whose weight is over a predefined threshold,
- or adding a fix number of goals, for example the k goals with the highest $pw(g_i)$ weight.

In the experiments reported in this paper we used the second technique. After a few tests, the number of goals to be added to the original query was set to 10.

# 6. Experimental Analysis

We decided to compare the two techniques (PPF and PRF) to see if one was better than the other from an experimental KE point of view. The motivation behind this analysis lies in the results of a previous investigation into a comparison of the results of the two previously described techniques [3]. In that study we discovered, given the best possible parameters settings, PPF performed as well as PRF, but extracted different sets of objects. The experimentation reported in this paper is more accurate, uses larger data sets, and aims at testing the possibility of combining the two techniques. We chosed to perform our analysis following the classical KE experimental methodology, using KE effectiveness measures and KE test collections. The main measures used in KE are Pertinent-Extracted and precision.

*Object-Pertinent-Extracted* (*ope*) is the sub-set of all objects in the collections that are pertinent to a query and that are actually extracted:

$$ope = \frac{pertinent\ and\ extracted\ objects}{pertinent\ objects} \quad (12)$$

*Precision* (PR) is the sub-set of the extracted objects that are also pertinent to the query:

$$pr = \frac{pertinent\ and\ extracted\ objects}{extracted\ objects} \quad (13)$$

In the following sections we present the results using pertinent-extracted and precision graphs (PR/PE graphs). These graphs are obtained by depicting the precision figures at standard levels of pertinent-extracted.

## 6.1 Experimental and Evaluation Settings

The object collections chosen for the investigation are the *Word processor*, and the *telephone's application* test collections. The main characteristics of these two test collections are summarised in Table 1. These are classical KE test collections used by our researchers in the field and described in [3, 4].

At the current stage, our research was not concerned with efficiency issues, and we decided that size was not so important. There are still many open issues in the application of SN to KE, and the problem of scaling the results is one of the major ones. We are aware of the fact that the collections used in our experiments can be considered small from the KE point of view.

## 6.2 Comparing Probabilistic Relevance and Possibilistic Pertinence Feedback

Since our first comparative look at PRF and PPF we noticed that the set of goals added to the initial query was very different. In other words, taking the same initial query and feeding back the same objects to the PPF and PRF we obtained different sets of goals (and different weights for the same goals) to be added to the query. In a test over 25 queries randomly chosen from the set of queries of our *Word processor* experimental collection we found that the difference between the two sets of m (m = 10) goals to be added to the query was on average 28%, i.e., at least 3 goals where different. The difference in weights between the same goals present in both sets was on average 87%. Analogous differences were found using the other test collections.

We compared the performance of PRF and PPF using training sets of different sizes. Figures 1 and 2 show graphically how the performance of PPF and PRF increases when the device is given an increasing number of pertinent objects.

The results reported in the tables refers to the *Word processor* collection, similar results were obtained using the *Telephone's application* collection. This shows what we expected: both PPF and PRF act like pattern recognition devices and the more information they receive the more they can discriminate between patterns of pertinent and not pertinent objects. The performance has been evaluated averaging over all queries in the pertinence assessment at different values of the number of pertinent objects given as feedback. The graphs show that the PPF increases more rapidly in performance than the PRF. This is due to the better characteristics of non linear discrimination of PPF, that enables it to separate better the two sets of pertinent and non-pertinent objects.

From the KE point of view, however, the performance of PRF are better than those of PPF. PRF is more effective at lower levels of training. This makes PRF more useful in applications where the percentage of pertinent objects versus the total number of object used in the pertinence feedback is usually low.

## 7. Limitations

In this paper we present a study of the effectiveness of possibilistic and probabilistic pertinence feedback on novice user's queries processing. We have reported some results in the previous sections of an analysis of the effects of the use of possibility and probability theories in user's queries on the effectiveness of an KE system. We believe the work presented here to be complete since it uses a larger number of query sets, a larger collection of objects, and a more classical KE system that was not tuned to the test collection used.

Nevertheless, there are at least two important limitations to this study:

1. The collection of the queries used in the experimentation were not really representative of typical novice user's queries. However, it has been long recognised that novice user's query is mainly dependent upon the application domain and the KE environment.

2. Dictated user's query is considerably different from spontaneous one. We should expect spontaneous novice user's queries to have higher levels of precision and pertinent-extracted. Unfortunately, there is no set of spontaneous novice user's queries available for query processing experimentation and its construction is not an easy task.

## 8. Conclusions and futur works

In this paper the briefly summarised of the results of this investigation, demonstrate that a Pertinence Feedback device based on PPF is more effective that PRF for low level of training. A low level of training (i.e., a small number of objects indicated as pertinent by the user) is typical of real life KE applications. However, since at any level of training PPF and PRF identify different sets of goals to be added to those present in the original query, a combination of PPF and PRF could prove effective. The results in this direction reported in this paper seem to indicate that the combination is indeed effective. Nevertheless, more tests with larger collections of objects and with different algorithms for PPF and PRF are necessary to validate these results and fully exploit their usefulness.

| Data | Word processor | Telephone's application |
|---|---|---|
| Objects | 140 | 132 |
| queries | 25 | 52 |
| goals in objects | 50 | 31 |
| goals in query | 3 | 2 |
| average object length | 6 | 5 |
| average query length | 15 | 11 |

Table 1. Test collections data

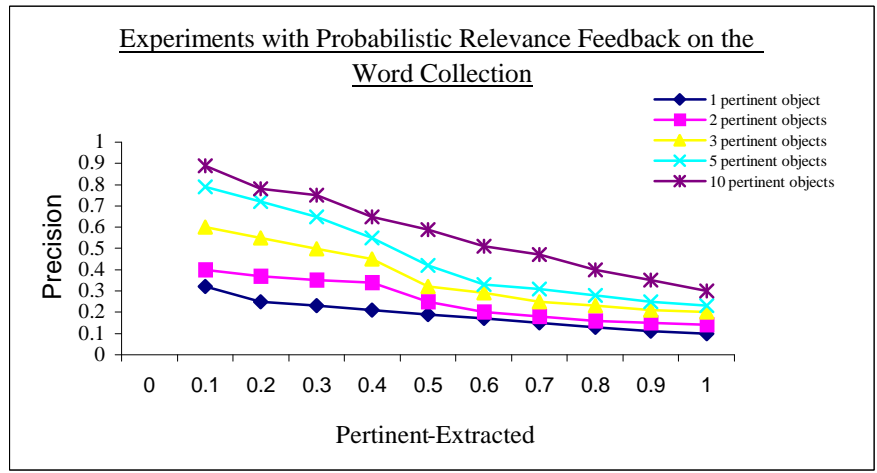

Fig. 1. Performance of Probabilistic Relevance Feedback.

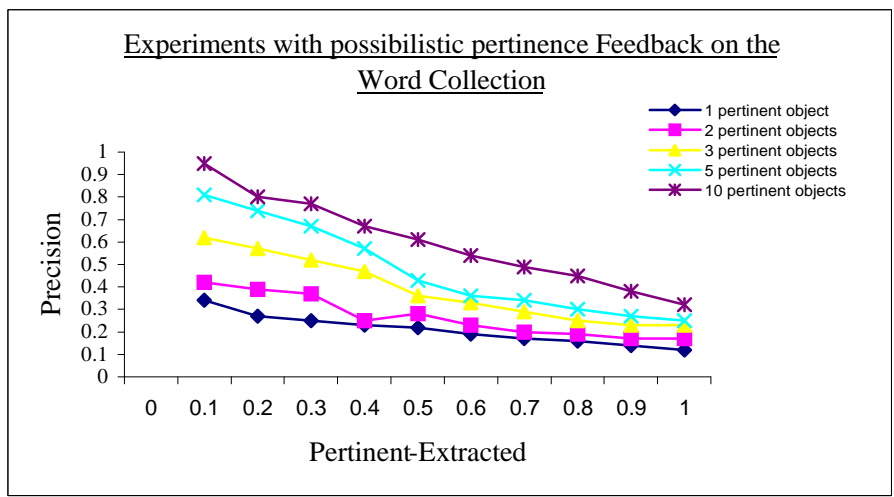

Fig. 2. Performance of Possibilistic Pertinence Feedback.